\documentclass[runningheads]{llncs}
\usepackage[strings]{underscore}
 
\usepackage{eccv}

\usepackage{eccvabbrv}

\usepackage{graphicx}
\usepackage{booktabs}

\usepackage[accsupp]{axessibility}  

\usepackage[pagebackref=true,breaklinks,colorlinks,allcolors=blue]{hyperref}

\usepackage{orcidlink}

\newcommand{\ob}{\textcolor[RGB]{0, 10, 215}}
\newcommand{\tb}{\textcolor[RGB]{0, 0, 0}}

\begin{document}

\title{\tb{The PanAf-SBR Dataset: Social Behaviour\\ Recognition for Wild Great Apes\vspace{-5pt}}} 

\titlerunning{PanAf-SBR}

\author{Maciej Braszczok\inst{1}\orcidlink{} \and
Otto Brookes\inst{1,2}\orcidlink{0000-0001-6865-1844} \and
Xiaoxuan Ma\inst{3}\orcidlink{0000-0003-0571-2659} \and
Federico Rossano\inst{4}\orcidlink{0000-0002-6544-7685} \and
Yixin Zhu\inst{5}\orcidlink{0000-0001-7024-1545} \and
Mimi Arandjelovic\inst{6}\orcidlink{0000-0001-8920-9684} \and
Hjalmar Kühl\inst{6, 7}\orcidlink{0000-0002-4440-9161} \and
Majid Mirmehdi\inst{1}\orcidlink{0000-0002-6478-1403} \and
Tilo Burghardt\inst{1}\orcidlink{0000-0002-8506-012X}
}

\authorrunning{M.~Braszczok et al.}

\institute{
University of Bristol, School of Computer Science, Bristol, UK 
\and
Wild Chimpanzee Foundation, Leipzig, Germany
\and
Carnegie Mellon University, Robotics Institute, Pennsylvania. USA
\and
University of California, Department of Cognitive Science, San Diego. USA
\and
Peking University, School of Computer Science, Beijing, China.
\and
Max Planck Institute for Evolutionary Anthropology, Leipzig, Germany
\and
Senckenberg Museum of Natural History, Görlitz, Germany
}

\maketitle

\begin{abstract}
\ob{\tb{Behavioural shifts in wild great ape populations, particularly the breakdown of social structures, can serve as an early indicator of population decline. Automating the detection of behaviours indicative of these shifts is therefore a critical task for conservation. Several valuable datasets have recently been introduced for the automated recognition of great ape behaviour, yet few include fine-grained social behaviour annotations, and those that do are captured either in captive settings or via aerial platforms such as UAVs. We address this gap by introducing PanAf-SBR, the first wild great ape camera trap dataset annotated with social behaviours. PanAf-SBR extends PanAf500 with 100 additional videos covering 36,063 frames. These come with 81,096 annotations including bounding boxes, segmentation masks, intra-video identities, and seven social behaviour classes defined under the action giver and receiver convention of ChimpACT. We use this data together with the AlphaChimp architecture to establish the first benchmarks for fine-grained social behaviour recognition in wild great apes from camera trap footage. We further conduct bidirectional transfer learning experiments between PanAf-SBR and the captive ChimpACT dataset, finding that cross-dataset pre-training is highly beneficial for specific classes rather than of uniform benefit. Finally, we examine the role of background context by inverting the segmentation masks to suppress non-ape pixels.}}
\keywords{\tb{Wild great apes \and Animal biometrics \and Social behaviour recognition \and Camera trap videos \and Transfer learning \and Computer Vision}\vspace{-8pt}}
\end{abstract}

\section{Introduction}
\label{sec:intro}\vspace{-4pt}


\ob{\tb{\textbf{Task \& Motivation}. All great ape species are currently classified as endangered or critically endangered by the IUCN~\cite{IUCN}, with several populations in steep decline. Behavioural shifts in wild populations -- particularly the breakdown of social structures -- can serve as early indicators of population collapse~\cite{kuhl2019human}. Beyond conservation, great ape social behaviour is also of long-standing interest to evolutionary anthropology, offering a comparative window onto the origins of human sociality, cooperation, and culture~\cite{boesch1996emergence,whiten2017culture}. Recognising the social behaviour of great apes from sensor data, such as camera trap footage, is therefore a critical task both for conservation and for the study of social evolution. However, the scale of footage collected by camera traps far outpaces the capacity of traditional methods to analyse it, and manual annotation remains a resource- and time-intensive process. Moreover, manual annotation is prone to human bias and to drift over time~\cite{fuchs2025forest}. Recent advances in spatio-temporal deep learning models offer a route to automating the recognition of ape social behaviour~\cite{zhuang2020comprehensive,vogg2025computer}, accelerating both ethological research and conservation efforts.}}

\begin{figure}[t]
\centering\vspace{-9pt}
\includegraphics[width=1.0\textwidth]
{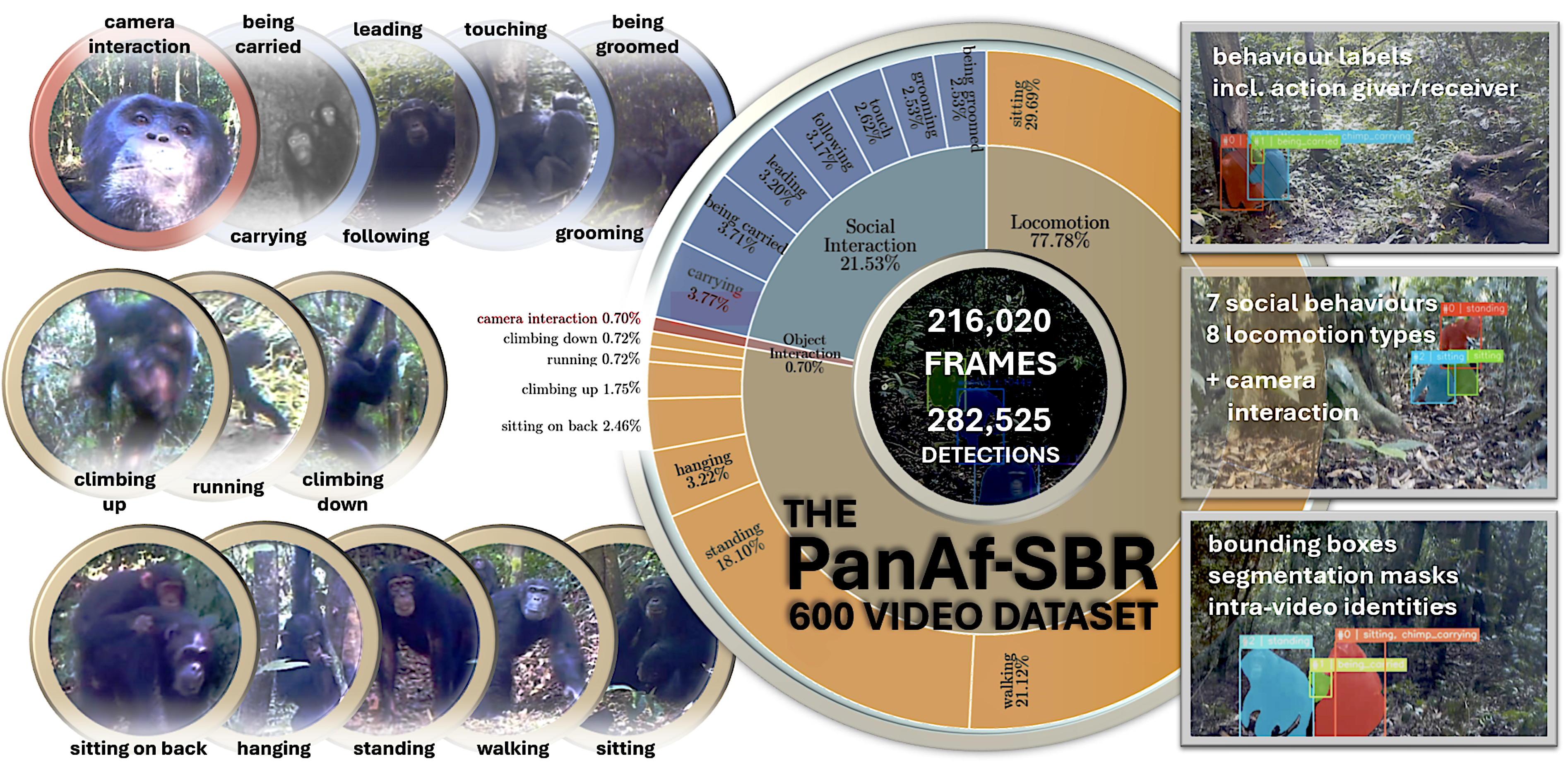}\vspace{-2pt}
\caption{\tb{\textbf{Overview of the PanAf-SBR Dataset.} The summary illustrates the dataset and its behavioural classes including seven social behaviours defined under the action giver and receiver convention of ChimpACT. Class example visuals are shown on the left clockwise relating to the statistics on the right. All social behaviours are defined according to the ethogram used in~\cite{ma2024alphachimp}, except for the \textit{leading} and \textit{following}, which are defined independently. The giver and receiver individuals are labelled as such and localised separately in the dataset with their own bounding box or segmentation mask.}
\vspace{-12pt}}
\label{fig:montage_grid_3}
\end{figure}

\ob{\tb{\textbf{State-of-the-art \& Limitations}. Several high-impact datasets and methods have recently been introduced to automate ape behaviour recognition, spanning both captive settings, where rich ethograms and pose annotations support fine-grained classification~\cite{Bala2020, Ma2023ChimpACT, Marks2022, ma2024alphachimp}, and the wild, where large-scale camera trap and UAV corpora and pre-training strategies have driven rapid gains in recognition accuracy~\cite{brookes2024panaf20k, mueller2026privi, Jing2026EthoCLIP}. Critically, however, the social dimension of behaviour remains largely unmodelled in the wild and particularly for camera traps. Captive corpora such as ChimpACT~\cite{Ma2023ChimpACT} offer exactly this role-differentiated social annotation, but captive housing alters group composition and removes the ecological pressures that shape wild social dynamics. Conversely, existing wild datasets~\cite{brookes2024panaf20k, brookes2025fgbg} do include social behaviours but their annotations exist only at video level or without role labels, since densely annotating multi-actor social roles in camera trap footage is far more costly than labelling individual actions for entire recordings. The result is a gap at the intersection of the two: no dataset combines the ecological authenticity of wild footage with the role-differentiated social annotation needed to model detailed social dynamics.}}

\ob{\tb{\textbf{Contribution}. In this work, we address this gap by extending the role-oriented social behaviour ethogrammatic approach introduced in ChimpACT~\cite{Ma2023ChimpACT} to wild camera trap footage from the PanAf datasets. Our contributions are as follows: \textbf{\textit{(i)}} we present PanAf-SBR, an extension of PanAf500 with 100 additional camera trap videos, 36,063 frames, and 81,096 annotations covering localisation, intra-video identity, and social behaviour. Each video retains one of the original nine PanAf500 behaviour classes and is further labelled with one or more of seven newly introduced social behaviour categories; localisation is annotated with both segmentation masks and bounding boxes, each linked to multi-label behaviour and intra-video identity labels; \textbf{\textit{(ii)}} we train AlphaChimp~\cite{ma2024alphachimp}, a network developed for end-to-end detection, tracking, and recognition of social interactions, on the dataset; \textbf{\textit{(iii)}} we conduct transfer learning experiments assessing the extent to which pre-training on captive chimpanzee footage benefits recognition in the wild, and vice versa and; \textbf{\textit{(iv)}} we investigate background dependencies by training and evaluating on foreground-only clips derived from inverted segmentation masks -- the first such analysis for great ape social behaviour recognition.}}

\section{Related Work}
\label{sec:related_work}\vspace{-8pt}

\ob{\tb{\textbf{Ape Behaviour Datasets}. Early primate behaviour datasets targeted neuroscience and ethology labs, where controlled imaging enabled rich annotation. OpenMonkeyStudio~\cite{Bala2020} used a 62-camera array to produce 3D skeletons for rhesus macaques. ChimpACT~\cite{Ma2023ChimpACT} provides 163 zoo-housed chimpanzee videos with detection, re-identification, pose, and a 23-class ethogram that uniquely distinguishes social roles (e.g.\ grooming vs.\ grooming-received). ChimpBehave~\cite{Fuchs2025ChimpBehave} adds 1,362 annotated segments with locomotion classes aligned to wild benchmarks. While valuable, these datasets lack in-situ footage from ecological sensors, limiting their suitability for studying natural behaviour. The PanAf20K~\cite{brookes2024panaf20k} is a large manually-curated wild behaviour dataset, containing 19,973 camera trap videos of chimpanzees and gorillas from 18 sites, with a 500-video subset densely annotated across 9 action classes without social behaviours. PanAf-FGBG~\cite{brookes2025fgbg} includes the social interactions \textit{grooming}, \textit{playing}, and \textit{aggression} but annotations exist only at the video level. BaboonLand~\cite{duporge2024baboonland} provides UAV drone footage of three wild baboon troops across 12 behaviours including social behaviours such as grooming. PriVi~\cite{mueller2026privi} instead targets pre-training scale over annotation, including 424 hours of primate footage pooled from 11 research sources and curated web video, surpassing all of the above in raw duration but providing no manually annotated behaviour labels of its own. EthoCLIP~\cite{Jing2026EthoCLIP} introduces AnimalBand, which combines 74K+ videos from major existing datasets standardised against the Neuro Behaviour Ontology -- larger still in video count, but spanning species generally rather than wild primates specifically. Animal Kingdom~\cite{ng2022animal} and MammalNet~\cite{chen2023mammalnet} offer greater taxonomic breadth (850 and 173 species) but draw mainly on curated footage. To date, no wild camera trap dataset offers the role-differentiated social annotation available in captive corpora such as ChimpACT.}}

\ob{\tb{\textbf{Ape Behaviour Recognition Methods}. In captive settings, pipelines increasingly unify detection, pose, and behaviour recognition: SIPEC~\cite{Marks2022} replaces explicit pose estimation with end-to-end classification, and AlphaChimp~\cite{ma2024alphachimp} performs detection, tracking, and behaviour recognition jointly via a DETR-style, Video-Swin-based framework, particularly effective for resolving multi-actor social behaviour. In the wild, early systems adapted human action recognition architectures~\cite{sakib2020visual,bain2021automated,brookes2023triple, brookes2024chimpvlm}, while ASBAR~\cite{Fuchs2023} showed skeleton-based recognition can match video-based methods on PanAf500 with far less input data. Pre-training V-JEPA~\cite{bardes2024revisiting} on PriVi~\cite{mueller2026privi} reaches 87.2\% top-1 accuracy on PanAf500 with a frozen classifier and no new behavioural labels, and EthoCLIP~\cite{Jing2026EthoCLIP} injects AnimalBand's ontology into a CLIP-style contrastive objective to improve recognition of rare or compositional behaviours.}}\vspace{-5pt}

\begin{figure}[t]
\centering
\includegraphics[width=290pt]{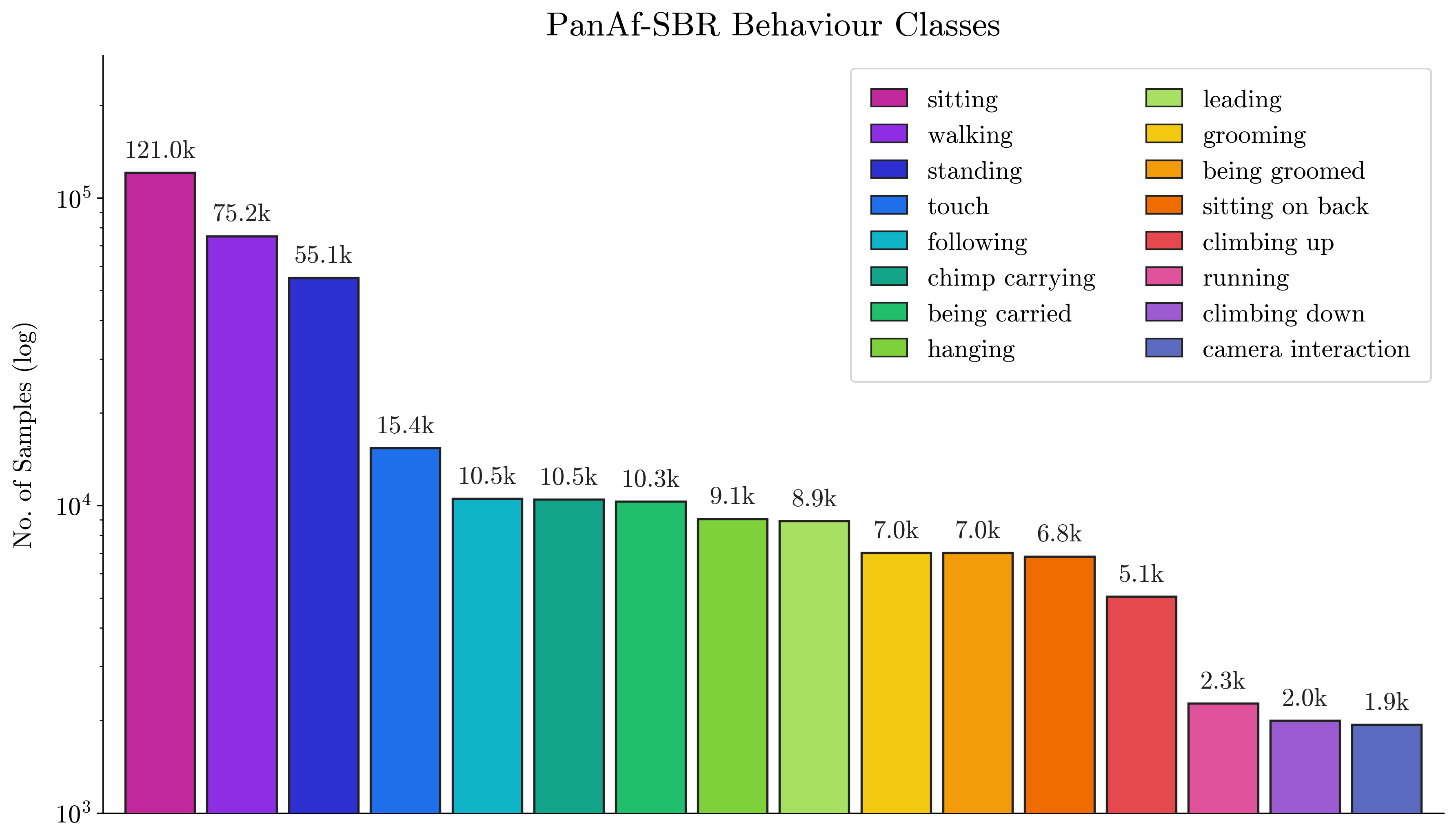}\vspace{-5pt}
\caption{\tb{\textbf{Distribution of the PanAf-SBR Behaviour Annotations}. The number of samples for each behaviour category is shown. The dataset follows a typically long-tailed distribution with many of the social behaviours falling within the middle.}}\vspace{-12pt}
\label{fig:PanAf-SBR_dist}
\end{figure}

\section{The PanAf-SBR Dataset}
\label{sec:dataset}\vspace{-7pt}

\subsection{Overview \& Statistics}\vspace{-4pt}

\ob{\tb{PanAf-SBR is a wild great ape camera trap dataset annotated for both locomotion and social behaviour recognition. It extends PanAf500 by 100 new videos selected for the presence of social behaviours, bringing the total to 600 videos, 216,020 frames, and 282,525 individual ape detections. Relative to PanAf500, this represents a $\sim$20\% increase in videos and frames, and a 40.2\% increase in detections; full split-level counts for both datasets are given in Table~\ref{tab:PanAf-SBRstatistics}. Each detection is paired with at least one behaviour label. Detections comprise both bounding boxes and segmentation masks. The dataset is partitioned into 480 training, 30 validation, and 90 test videos, with camera locations held disjoint across splits to prevent localisation bias. The per-class detection distribution, shown in Figure~\ref{fig:PanAf-SBR_dist}, is heavily long-tailed, consistent with the class imbalance characteristic of wildlife behaviour datasets~\cite{brookes2024panaf20k}. Among locomotion classes, \textit{sitting} and \textit{walking} are the most frequently observed. Social behaviours collectively account for approximately 21.53\% of all detections, with \textit{leading} and \textit{following} being the rarest classes in the dataset.}}

\ob{\tb{\textbf{Ethical Considerations.} PanAf-SBR builds exclusively on footage already publicly released as part of PanAf20k~\cite{brookes2024panaf20k}, which was itself gathered under ethical oversight as part of the Pan African Programme: The Cultured Chimpanzee~\cite{PanAf}. To protect wild chimpanzee populations from threats such as poaching and habitat disturbance, all information that could be used to infer the location of the animals has been withheld or anonymised. Specifically, country names, research site names, and precise geospatial coordinates are not released. No individual ape identities are disclosed. These measures are consistent with those in PanAf20k following established best practices for ethical release of wildlife data~\cite{brookes2024panaf20k}.}}


\begin{table}[t]
  \scriptsize
  \centering
  \caption{
    \tb{\textbf{Frame and Detection Counts in PanAf-SBR against PanAf500.} \textmd{PanAf-SBR extends PanAf500 with 100 additional videos selected for presence of social behaviours, increasing the total video and frame count by 20\% and detection count by 40.2\%. Each detection is a bounding box paired with at least one behaviour label.}}\vspace{-3pt}
  }
  \label{tab:PanAf-SBRstatistics}
  \begin{tabular}{@{}lrrrrrr@{}}
  \toprule
  & \multicolumn{3}{c}{\textbf{PanAf-SBR}} & \multicolumn{3}{c}{\textbf{PanAf500}} \\
  \cmidrule(lr){2-4} \cmidrule(lr){5-7}
  \textbf{Split} & \textbf{Videos} & \textbf{Frames} & \textbf{Detections} & \textbf{Videos} & \textbf{Frames} & \textbf{Detections} \\
  \midrule
  Train      & 480 & 172{,}831 & 218{,}132 & 400 & 143{,}959 & 154{,}713 \\
  Validation &  30 &  10{,}800 &  19{,}360 &  25 &   9{,}000 &  14{,}744 \\
  Test       &  90 &  32{,}389 &  45{,}033 &  75 &  26{,}998 &  31{,}973 \\
  \midrule
  Total      & 600 & 216{,}020 & 282{,}525 & 500 & 179{,}957 & 201{,}430 \\
  \bottomrule
  \end{tabular}\vspace{-9pt}
\end{table}

\vspace{-12pt}
\subsection{Data Sourcing and Preparation}\vspace{-6pt}
\label{sec:dataset:source}

\ob{\tb{PanAf-SBR builds on PanAf20k~\cite{brookes2024panaf20k}, which comprises approximately 20,000 15-second camera trap videos from sites across western and central Africa, each accompanied by a video-level multi-label behaviour annotation; labels are neither attributed to individual apes nor temporally localised. PanAf500~\cite{sakib2020visual, brookes2024panaf20k} is a curated subset of 500 such videos, identical in all respects except that each is re-annotated at 24 fps with per-frame bounding boxes and a single behaviour label per individual. The dataset presented here extends the PanAf500 by 100 videos sampled from PanAf20k and selected for the presence of social behaviours. Candidates were identified by filtering PanAf20k annotations for the labels \textit{social interaction}, \textit{chimp carry}, and \textit{grooming}, yielding approximately 2,000 videos. Each was then manually inspected for clearly visible, sustained social interaction between two or more individuals, from which~100 were selected for annotation.}}

\begin{figure}[t]
    \centering
    \includegraphics[width=0.49\textwidth]{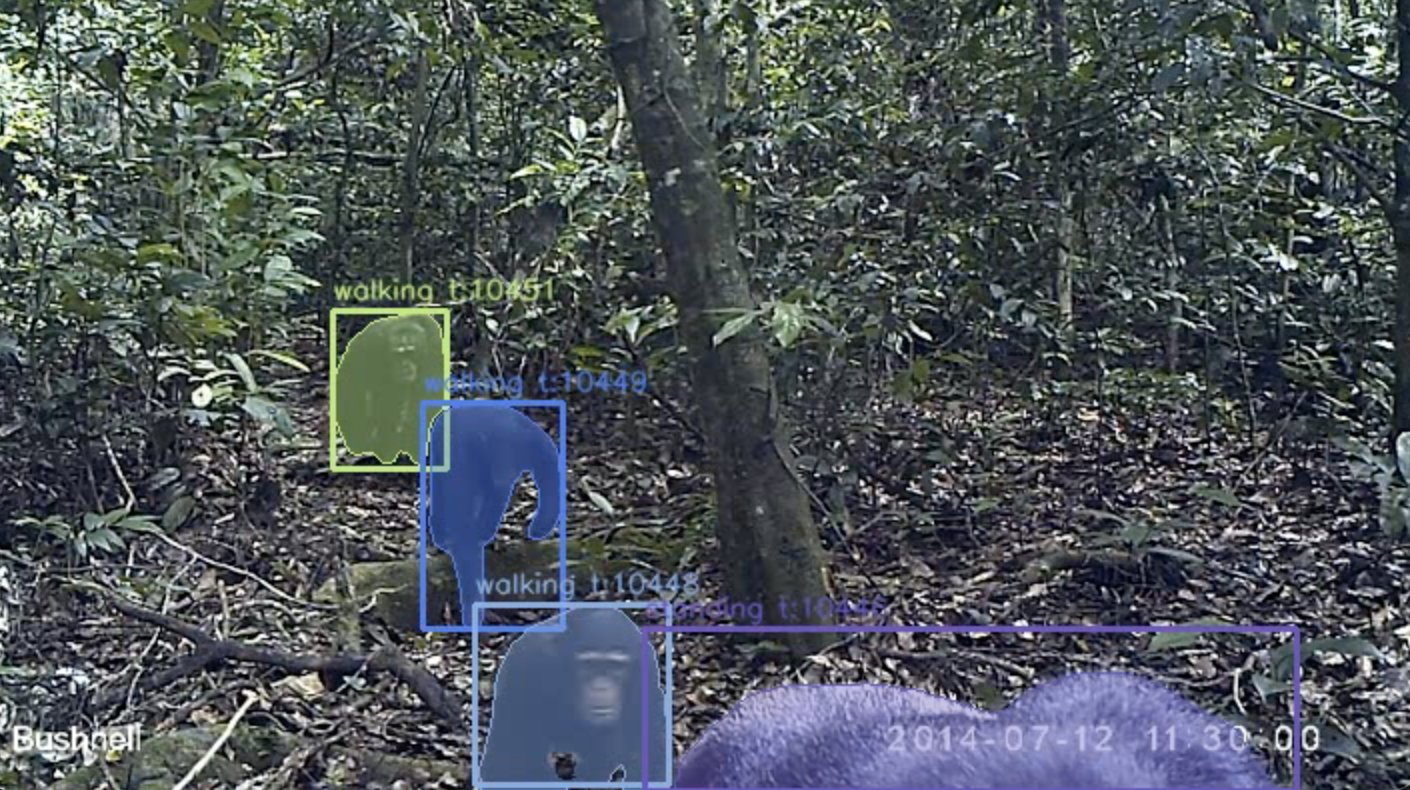}\hfill
    \includegraphics[width=0.49\textwidth]{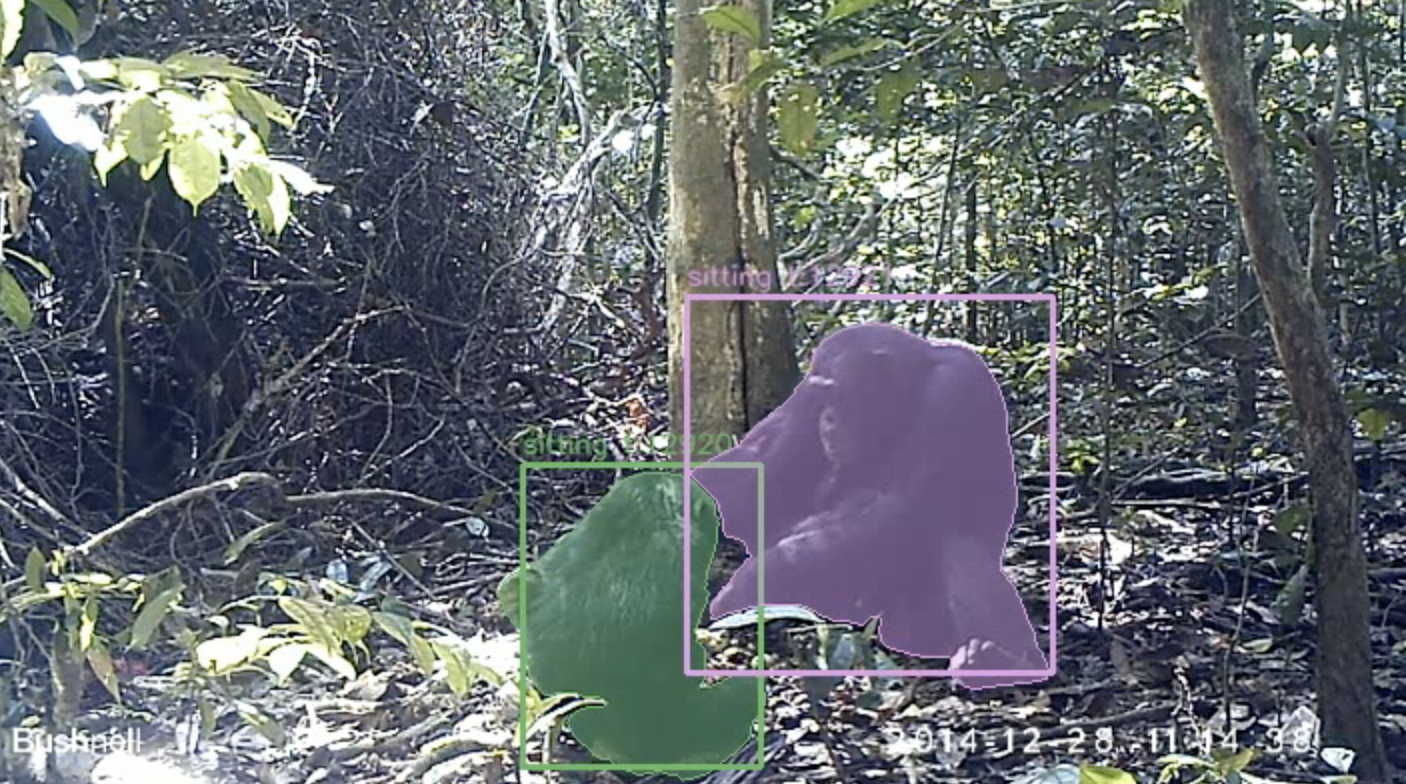}\\[0.5em]
    \includegraphics[width=0.49\textwidth]{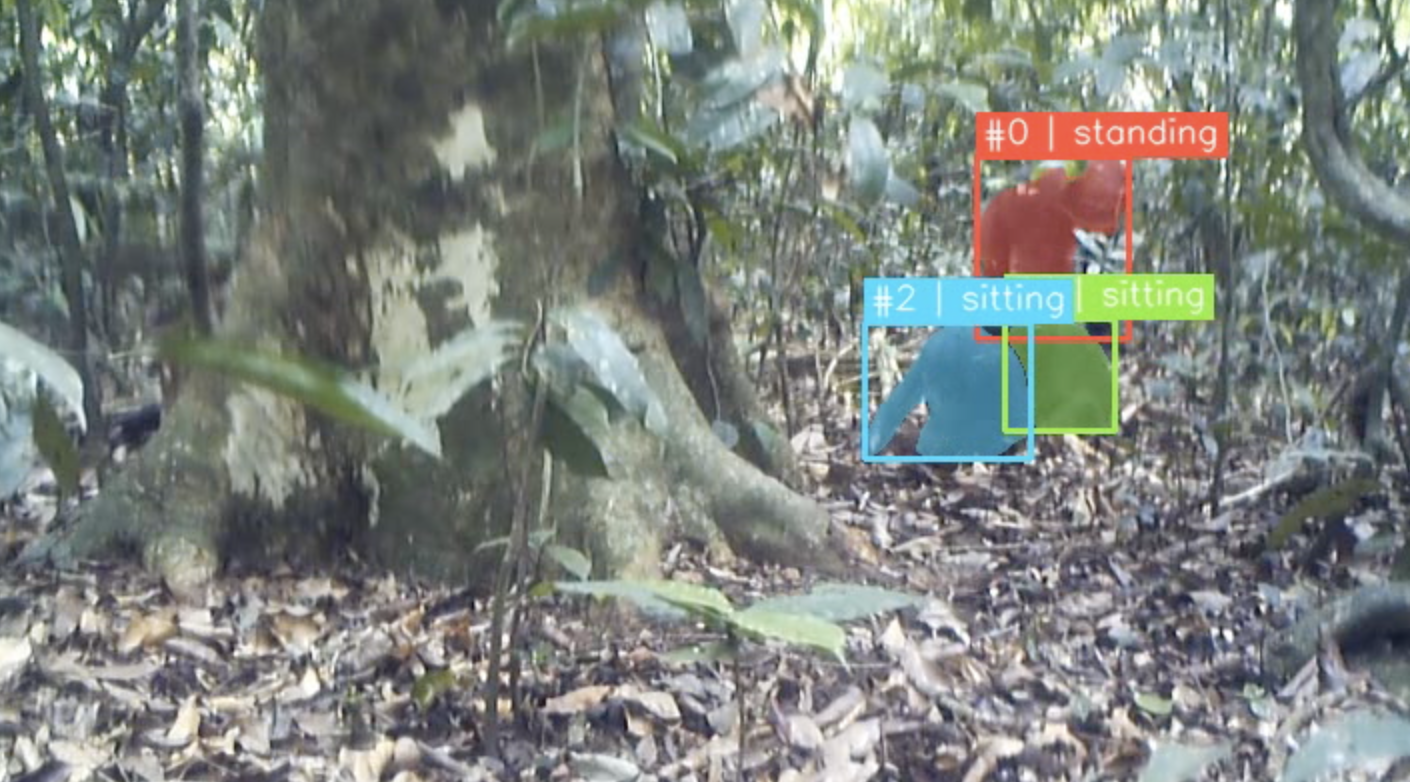}\hfill
    \includegraphics[width=0.49\textwidth]{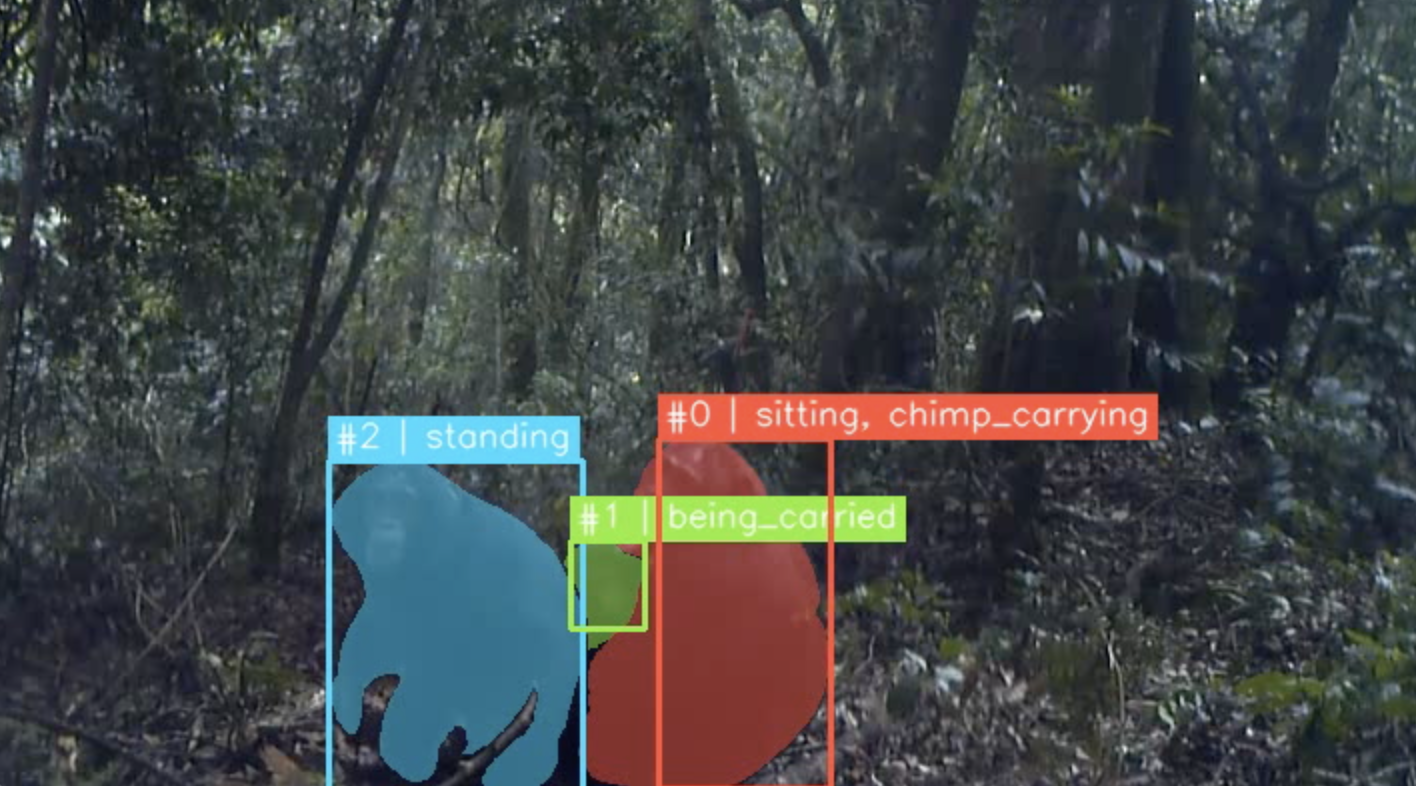}\\[0.5em]\vspace{-8pt}
    \caption{\tb{\textbf{Data Annotations.} Segmentation masks and recovered bounding boxes by the SA-FARI fine-tuned SAM3 model are shown with integers and colours encoding intra-video identity of localised animals. Frame-wise behaviour is shown as text. Bottom right image shows locomotive and social behaviour (i.e., sitting + chimp carrying).}}\vspace{-16pt}
    \label{fig:correctaaa}
\end{figure}

\vspace{-12pt}
\subsection{Data Annotation}\vspace{-6pt}
\label{sec:dataset:annotation}

\ob{\tb{\textbf{Localisation Annotation}. Localisations for the 100 new videos were generated by applying SAM~3 fine-tuned on SA-FARI~\cite{safari}~(see Figure~\ref{fig:correctaaa}), with the prompt ``chimpanzee''. The resulting bounding boxes, recovered from the outputted segmentation masks, were then exhaustively reviewed by a single annotator. Common failure modes -- duplicate detections of the same individual and identity switches across frames -- were corrected using purpose-built scripts for removing detections by individual ID and frame range. Short gaps where an individual was clearly visible but undetected were closed by linear interpolation between the surrounding bounding boxes; longer gaps where the individual changed direction or speed were first anchored with manually placed boxes before interpolation. Finally, to obtain high-fidelity segmentation masks, SAM~3 was re-prompted using the corrected bounding boxes and the outputs reviewed manually.}}

\ob{\tb{The same pipeline was applied to the 500 PanAf500 videos. Although SA-FARI provides human-verified segmentation masks for 429 of these videos, direct reuse is precluded by two mismatches: SA-FARI annotations are sampled at 6~fps, whereas PanAf-SBR requires 24~fps, and the remaining 71 PanAf500 videos have no SA-FARI coverage. The SAM~3 pipeline was therefore applied to all 500 videos to generate a consistent set of 24~fps localisations across the full dataset.}}

\ob{\tb{\textbf{Behaviour Annotation}. PanAf500 annotates each individual with one of nine labels (\textit{walking}, \textit{running}, \textit{climbing_up}, \textit{climbing_down}, \textit{hanging}, \textit{standing}, \textit{sitting}, \textit{sitting_on_back}, and \textit{camera_interaction}). We extend this label set with seven new social behaviour labels, defined under the action giver and receiver convention of ChimpACT~\cite{ma2024alphachimp} (see Figure~\ref{fig:montage_grid_3}). Five of the seven labels: \textit{grooming}, \textit{being groomed}, \textit{chimp carrying}, \textit{being carried} and \textit{touch} are adopted from ChimpACT~\cite{ma2024alphachimp} to enable direct transfer learning experiments, and the first four also correspond to PanAf20k video level annotations~\cite{brookes2024panaf20k}: \textit{grooming} and \textit{chimp_carrying} that guided our 100-video selection. \textit{Leading} and \textit{following} are novel to this work, capturing coordinated travel and does not involve physical contact, unlike any of the other social classes. Behaviour annotations follow a multi-label convention: every individual present in a frame receives exactly one locomotion label, with those engaged in social interaction additionally assigned multiple distinct social labels; any pair of apes participates in at most one social behaviour at a time, but a single ape may simultaneously interact with several others in the frame, accumulating to multiple concurrent social labels. Both bounding boxes and behaviour labels are annotated at 24~fps, yielding up to 360 per-frame annotations per video.}}


\vspace{-12pt}
\section{Experiments}\vspace{-7pt}
\label{sec:experiments}

\subsection{Experimental Setup}\vspace{-4pt}

\begin{figure}[t]
\centering
\includegraphics[width=\textwidth]{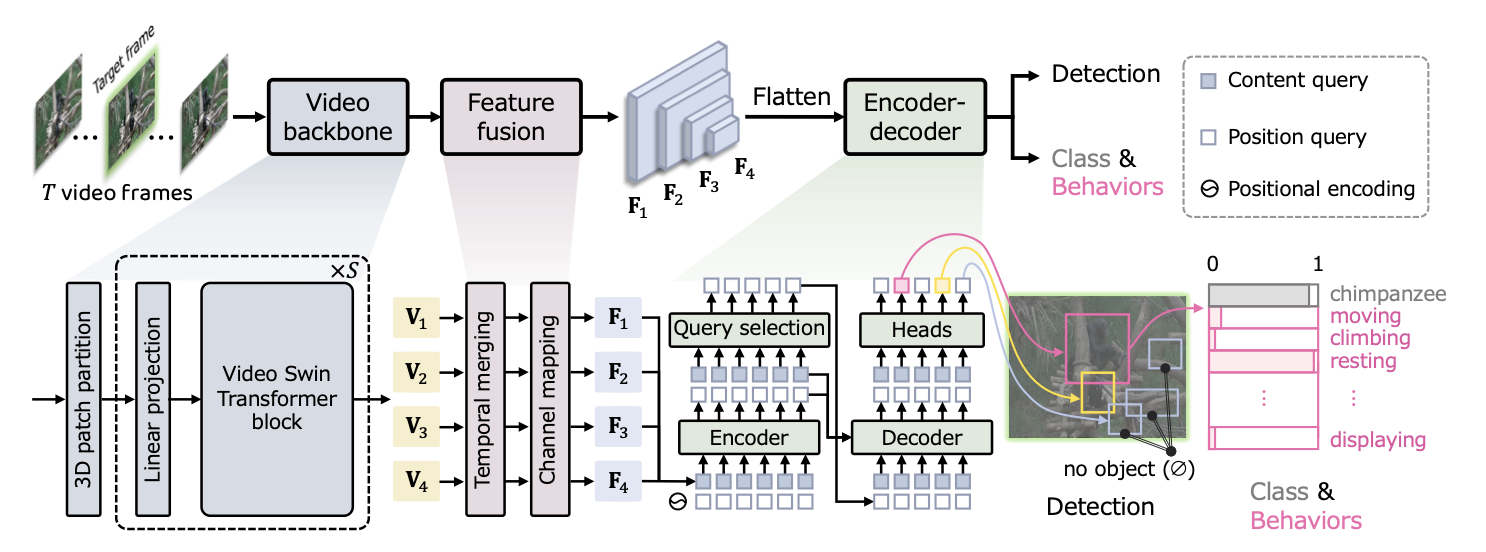}
\caption{\tb{\textbf{Overview of the Employed AlphaChimp Architecture.} A Video Swin Transformer backbone, a convolutional temporal fusion layer, and a DINO detection head with a multi-label classification branch are utilised to form the AlphaChimp pipeline. (Image: courtesy of~\cite{ma2024alphachimp}).}}\vspace{-12pt}
\label{fig:alphachimp}
\end{figure}

\ob{\tb{All experiments use AlphaChimp~\cite{ma2024alphachimp} as the base model. AlphaChimp is an end-to-end framework that jointly localises individuals with bounding boxes, tracks them across frames, and predicts multiple behaviour classes. Its architecture comprises three components (illustrated in Figure~\ref{fig:alphachimp}): a Video Swin-L~\cite{liu2022video} backbone comprising four Swin Transformer blocks, which produces multi-scale spatio-temporal features at spatial strides of 4, 8, 16, and 32; a lightweight convolutional fusion module that collapses the temporal dimension onto a single target frame and projects all scales to a common channel dimension; and a DINO~\cite{zhang2022dino} detection head with 12 encoder and 12 decoder layers, four reference points in the deformable attention module, and an attached multi-label behaviour classification branch. We adapt the AlphaChimp architecture to PanAf-SBR by replacing the linear layer in the classification head to output 16 classes. All other architectural and optimisation hyperparameters follow the original defaults: input sequences of $T=8$ frames at a resolution of $576\times576$, a decoder query number of $Q=10$, and a batch size of 64; the temporal stride is set to 4 frames. The Video Swin-L backbone is initialised with ImageNet-21k weights, and the full detection model is further pre-trained on Object365~\cite{shao2019objects365} using weights provided by the AlphaChimp authors. All training runs were performed on $4\times$NVIDIA H100 GPUs.}}

\ob{\tb{\textbf{Evaluation.} All models are evaluated on the PanAf-SBR test split. We report precision ($\frac{TP}{TP+FP}$), recall ($\frac{TP}{TP+FN}$), and F1 score ($\frac{2PR}{P+R}$) at an intersection-over-union (IoU) threshold of 0.5, alongside average precision ($AP$) per class and mean average precision ($mAP$) aggregated across classes, following the evaluation protocol of the original AlphaChimp paper~\cite{ma2024alphachimp}. Results are reported separately for locomotion, social interaction, and object interaction classes, as well as aggregated across all classes.}}

\ob{\tb{\textbf{Mapping between PanAf-SBR and ChimpACT classes.} Although ChimpACT defines 23 behaviour classes against PanAf-SBR's 16, an intuitive mapping exists between the two ethograms, summarised in Table~\ref{tab:label_mapping}. The correspondence is many-to-one for locomotion: PanAf-SBR's seven fine-grained locomotion classes collapse onto three coarser ChimpACT categories, with \textit{walking} and \textit{running} both mapping to \textit{moving}; \textit{climbing_up}, \textit{climbing_down}, and \textit{hanging} all mapping to \textit{climbing}; and \textit{standing} and \textit{sitting} both mapping to \textit{resting}.}}

\ob{\tb{Five of the seven PanAf-SBR social classes have direct ChimpACT counterparts: \textit{grooming}, \textit{being_groomed} and \textit{being_carried} map one-to-one, while \textit{chimp_carrying} maps to ChimpACT's \textit{carrying} and \textit{touch} to \textit{touching}. The two novel PanAf-SBR classes, \textit{leading} and \textit{following}, have no ChimpACT equivalent. Finally, \textit{camera_interaction} maps to ChimpACT's \textit{solitary_object_playing}, the closest behavioural analogue for investigative object-directed behaviour.}}

\begin{table}[t]
  \scriptsize
  \centering
  \caption{
    \textbf{\tb{PanAf-SBR to ChimpACT Label Mapping.}}
    \textmd{\tb{Correspondence between PanAf-SBR and ChimpACT behaviour labels. PanAf-SBR locomotion classes map onto three coarser ChimpACT categories; five social classes have direct ChimpACT counterparts; \textit{leading} and \textit{following} have no ChimpACT equivalent.}}
  }
  \label{tab:label_mapping}
  \begin{tabular}{@{}lll@{}}
  \toprule
  \textbf{PanAf-SBR Label} & \textbf{ChimpACT Label} & \textbf{Category} \\ \midrule
  walking             & moving                    & Locomotion \\
  running             & moving                    & Locomotion \\
  climbing\_up        & climbing                  & Locomotion \\
  climbing\_down      & climbing                  & Locomotion \\
  hanging             & climbing                  & Locomotion \\
  standing            & resting                   & Locomotion \\
  sitting             & resting                   & Locomotion \\ \midrule
  grooming            & grooming                  & Social Interaction \\
  being\_groomed      & being\_groomed            & Social Interaction \\
  chimp\_carrying     & carrying                  & Social Interaction \\
  being\_carried      & being\_carried            & Social Interaction \\
  touch               & touching                  & Social Interaction \\
  leading             & ---                       & Social Interaction \\
  following           & ---                       & Social Interaction \\ \midrule
  camera\_interaction & solitary\_object\_playing & Object Interaction \\
  \bottomrule
  \end{tabular}
\end{table}

\subsection{Captive-to-Wild Experiments}
\label{sec:baseline_c2w}

\begin{figure}[t]
\centering
\includegraphics[width=290pt]{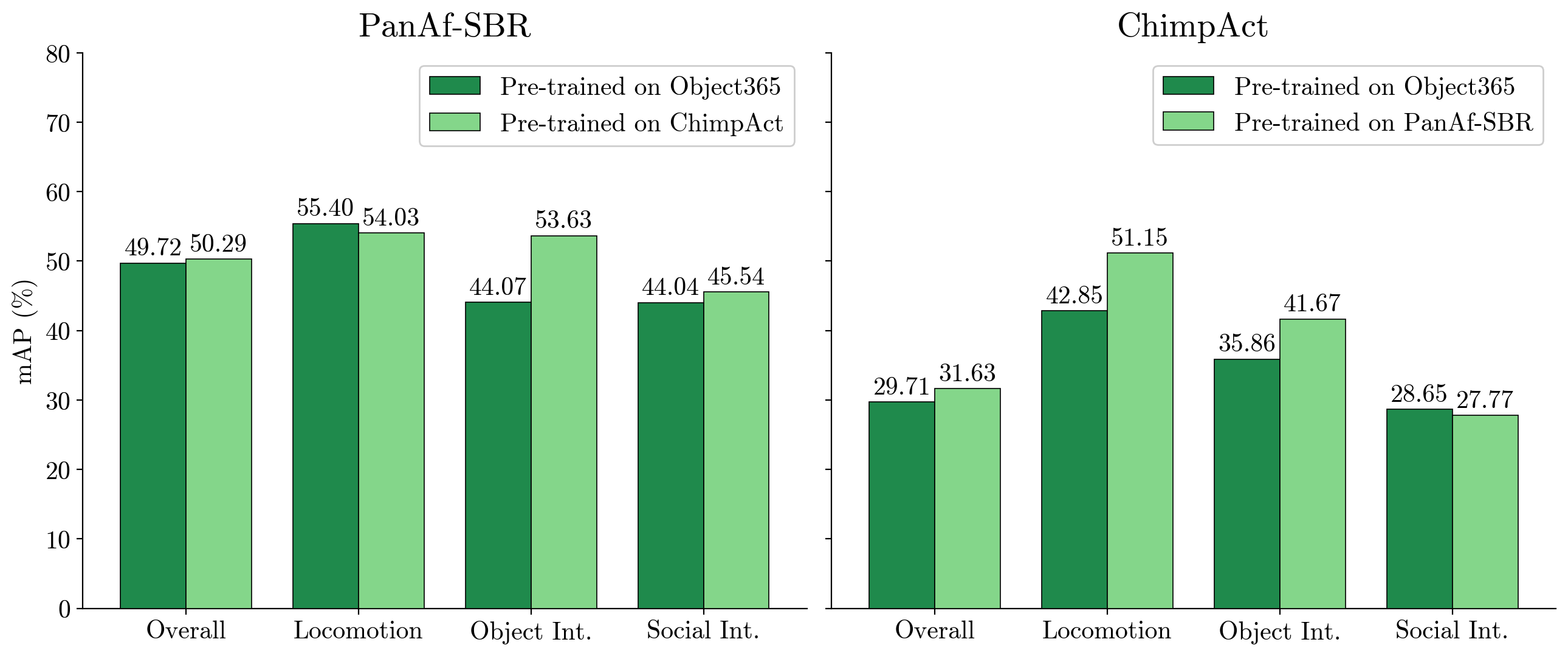}\vspace{-6pt}
  \caption{
    \tb{\textbf{Per-category AP at IoU 0.5 for Cross-Dataset Pre-Training.}
    \textmd{Results are grouped into locomotion, object interaction, and social interaction categories. Object interaction is the most consistent beneficiary of cross-dataset pre-training in both directions, while locomotion and social interaction transfer asymmetrically: wild-to-captive pre-training benefits locomotion substantially but the reverse direction does not hold, and social interaction shows only modest movement in either direction. All scores are AP (\%) at IoU and classification threshold of 0.5.} 
  }}\vspace{-13pt}
\label{fig:pretrain_ablation}
\end{figure}

\ob{\tb{\textbf{ChimpACT Pre-training on PanAf-SBR}. We first establish a baseline by training AlphaChimp directly on PanAf-SBR, then investigate whether pre-training on the captive dataset ChimpACT~\cite{ma2024alphachimp} is beneficial. As shown in Table~\ref{tab:baseline}, ChimpACT pre-training yields only a marginal improvement in mAP (+0.57\%, 49.72\% to 50.29\%), but substantially larger gains in recall (+19.24\%), precision (+6.13\%), and F1 (+13.19\%) at a fixed classification threshold of 0.5. At the category level, as shown in Fig.~\ref{fig:pretrain_ablation}, object interaction benefits most (+9.56\%), social interaction improves modestly (+1.50\%), while locomotion regresses slightly (-1.37\%). The improvement in object and social interaction is consistent with the focus of the ChimpACT dataset that was designed with a specific focus on interaction relative to the PanAf's focus on locomotive behaviours.}} 

\begin{table}[t]
  \scriptsize
  \centering
  \caption{
    \tb{\textbf{Effect of Cross-Dataset Pre-Training.}
    \textmd{Top: AlphaChimp trained and evaluated on PanAf-SBR, with and without ChimpACT pre-training. Bottom: AlphaChimp trained and evaluated on ChimpACT, with and without PanAf-SBR pre-training. The ChimpACT baseline mAP is taken from the original AlphaChimp paper~\cite{ma2024alphachimp}; precision, recall, and F1 are not reported for this configuration. All scores are at IoU threshold and classification threshold of 0.5.}
  }}\vspace{-4pt}
  \label{tab:baseline}
  \begin{tabular}{@{}llrrrr@{}}
  \toprule
  \textbf{Pre-train} & \textbf{Train} & \textbf{mAP} & \textbf{Precision} & \textbf{Recall} & \textbf{F1} \\ \midrule
  Object365                & PanAf-SBR       & 49.72 & 25.82 & 21.53 & 20.02 \\
  ChimpACT           & PanAf-SBR       & 50.29 & 31.95 & 40.77 & 33.21 \\
  Perf. ($\Delta$)           &         ---       & +0.57 & +6.13 & +19.24 & +13.19 \\ \midrule
  Object365                & ChimpACT       & 29.71 & 12.31   & 17.95   & 13.55   \\
  PanAf-SBR           & ChimpACT       & 31.63 & 13.55 & 14.45 & 12.79 \\
  Perf. ($\Delta$)           &         ---       & +1.92 & +1.23   & $-$3.51   & $-$0.76   \\
  \bottomrule
  \end{tabular}
\end{table}\vspace{-2pt}

\begin{figure}[t]
\centering
\includegraphics[width=\textwidth]{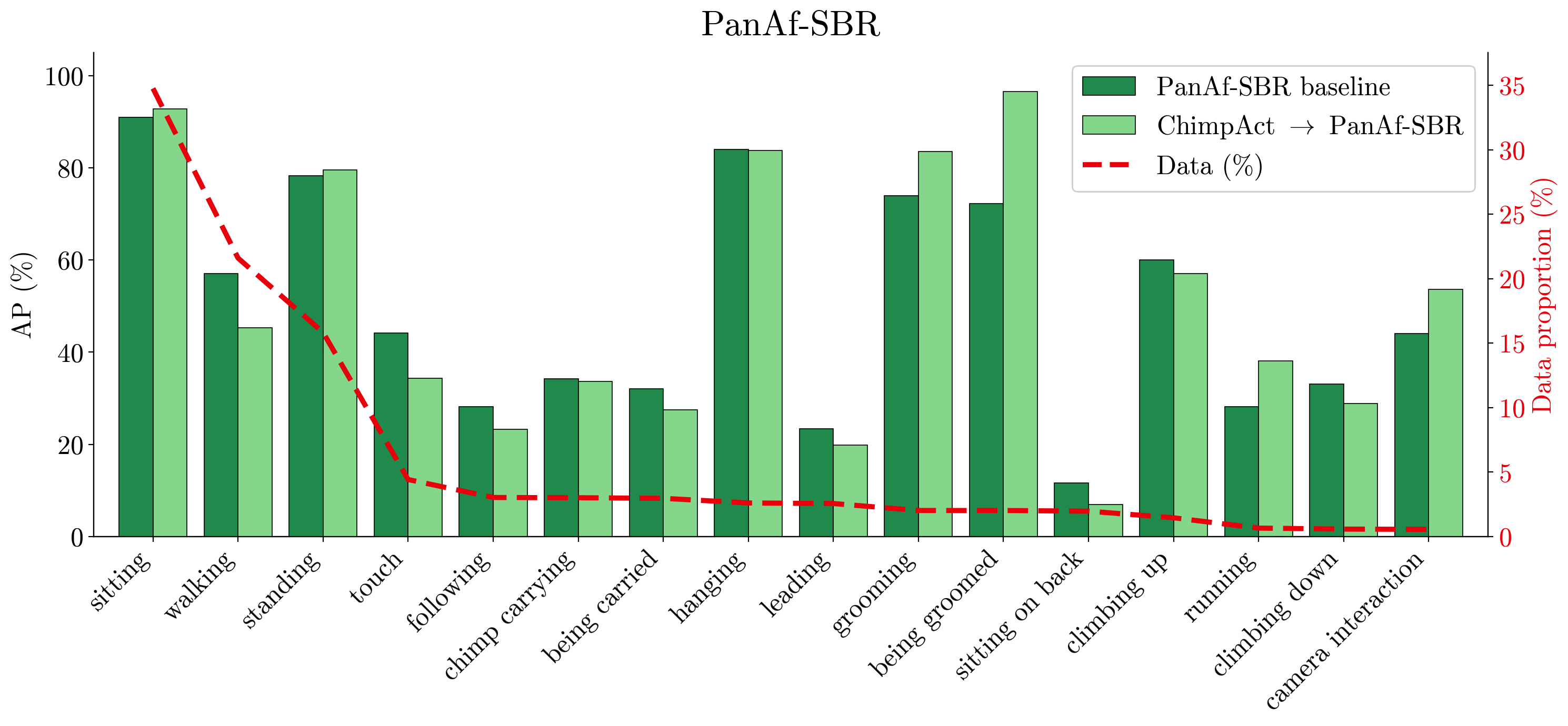}\vspace{-7pt}
\caption{\tb{\textbf{Per-Class AP on PanAf-SBR.} Shown is a comparison of the baseline model against the variant pre-trained on ChimpACT. Two social behaviours (\textit{grooming} and \textit{being\_groomed}) and one object interaction behaviour (\textit{camera\_interaction}) benefit from ChimpACT pre-training. The two locomotion behaviours \textit{walking} and \textit{running} diverge sharply: the former regresses under pre-training while the latter improves.} 
}
\label{fig:per_class1}
\end{figure}

\ob{\tb{\textbf{ChimpACT Pre-Training most benefits \emph{`grooming'}-related Classes.} Although ChimpACT pre-training improves social interaction on aggregate, the per-class results (see Fig.~\ref{fig:per_class1}) reveal that only two closely related social behaviours benefit: \textit{grooming} (73.96\% to 83.56\%, +9.60\%) and \textit{being_groomed} (72.25\% to 96.56\%, +24.31\%). \textit{camera_interaction} also benefits substantially (44.07\% to 53.63\%, +9.56\%). Notably, their ChimpACT counterparts — \textit{object interaction}, \textit{grooming}, and \textit{resting} — are among the most frequent classes in the ChimpACT training set, accounting for 30.5\%, 14.2\%, and 5.5\% of all training instances and ranking 1st, 2nd, and 8th respectively. However, class frequency alone does not explain the transfer outcomes: \textit{touching} accounts for 7.5\% of ChimpACT training instances (4th rank), yet its PanAf-SBR counterpart \textit{touch} regresses from 44.14\% to 34.35\% ($-$9.79\%), suggesting that frequency of occurrence in pre-training data is necessary but not sufficient for good transfer.}}

\ob{\tb{\textbf{Mapped Locomotion Classes have Heterogeneous Performance.} The \textit{resting} mapping is the most consistent, with both \textit{sitting} (90.99\% to 92.82\%, +1.83\%) and \textit{standing} (78.28\% to 79.49\%, +1.21\%) improving modestly — the only many-to-one locomotion mapping where all constituent classes benefit. The \textit{climbing} mapping uniformly shows reduced performance across \textit{climbing_up} (60.00\% to 57.07\%, $-$2.93\%), \textit{climbing_down} (33.12\% to 28.91\%, $-$4.21\%), and \textit{hanging} (83.94\% to 83.75\%, $-$0.19\%), suggesting the ChimpACT \textit{climbing} representation is incompatible with the corresponding wild behaviours. The \textit{moving} mapping is the most notable: \textit{running} improves by +9.90\% (28.15\% to 38.05\%) while \textit{walking} regresses by $-$11.79\% (57.04\% to 45.25\%) -- a divergence of 21.69\% between two classes sharing identical pre-training supervision.}}

\ob{\tb{\textbf{No Pre-Training Benefit for Social Classes with no ChimpACT Counterpart.} \textit{leading} (23.42\% to 19.89\%, $-$3.53\%) and \textit{following} (28.17\% to 23.28\%, $-$4.89\%) show reductions in performance of similar magnitudes, forming a tight cluster that is broadly consistent with \textit{chimp_carrying} (34.25\% to 33.60\%, $-$0.65\%) and \textit{being_carried} (32.09\% to 27.54\%, $-$4.55\%) -- suggesting a general pattern of moderate negative transfer across social classes that either lack a counterpart or whose counterpart is insufficiently represented in ChimpACT.}}

\subsection{Wild-to-Captive Experiments}
\label{sec:baseline_w2c}\vspace{-3pt}

\ob{\tb{\textbf{PanAf-SBR Pre-training on ChimpACT}. Additionally, we investigate the reverse direction -- the effect of PanAf-SBR pre-training on ChimpACT performance. As shown in Table~\ref{tab:baseline}, mAP on ChimpACT improves by +1.92\% (29.71\% to 31.63\%), with Fig~\ref{fig:pretrain_ablation} revealing a strong locomotion gain (+8.30\%) and a consistent object interaction improvement (+5.81\%). Social interaction, by contrast, is essentially unchanged ($-$0.88\%, 28.65\% to 27.77\%). Object interaction is the most consistent beneficiary of cross-dataset pre-training in both directions (+9.56 and +5.81\%), suggesting that object-directed behaviours generalise well across the captive-wild divide. Locomotion transfer is asymmetric: pre-training on wild data benefits captive locomotion recognition substantially (+8.30\%), while the reverse direction shows a small regression ($-$1.37\%). Social interaction transfer, by contrast, is consistently modest in both directions, with neither configuration producing a particularly meaningful change.}}

\ob{\tb{\textbf{Locomotion Classes Benefit from PanAf-SBR Pre-Training.} \textit{sleeping} improves substantially (+18.80\%), followed by \textit{climbing} (+8.38\%) and \textit{moving} (+6.26\%); only \textit{resting} is essentially unchanged (-0.26\%). This is consistent with the strong aggregate locomotion gain reported in Fig.~\ref{fig:pretrain_ablation} (+8.30\%) and suggests that wild locomotion footage, despite its visual differences from the captive setting, provides a useful prior for ChimpACT movement, climbing, and rest states.}}

\begin{figure}[t]
\centering
\includegraphics[width=\textwidth]{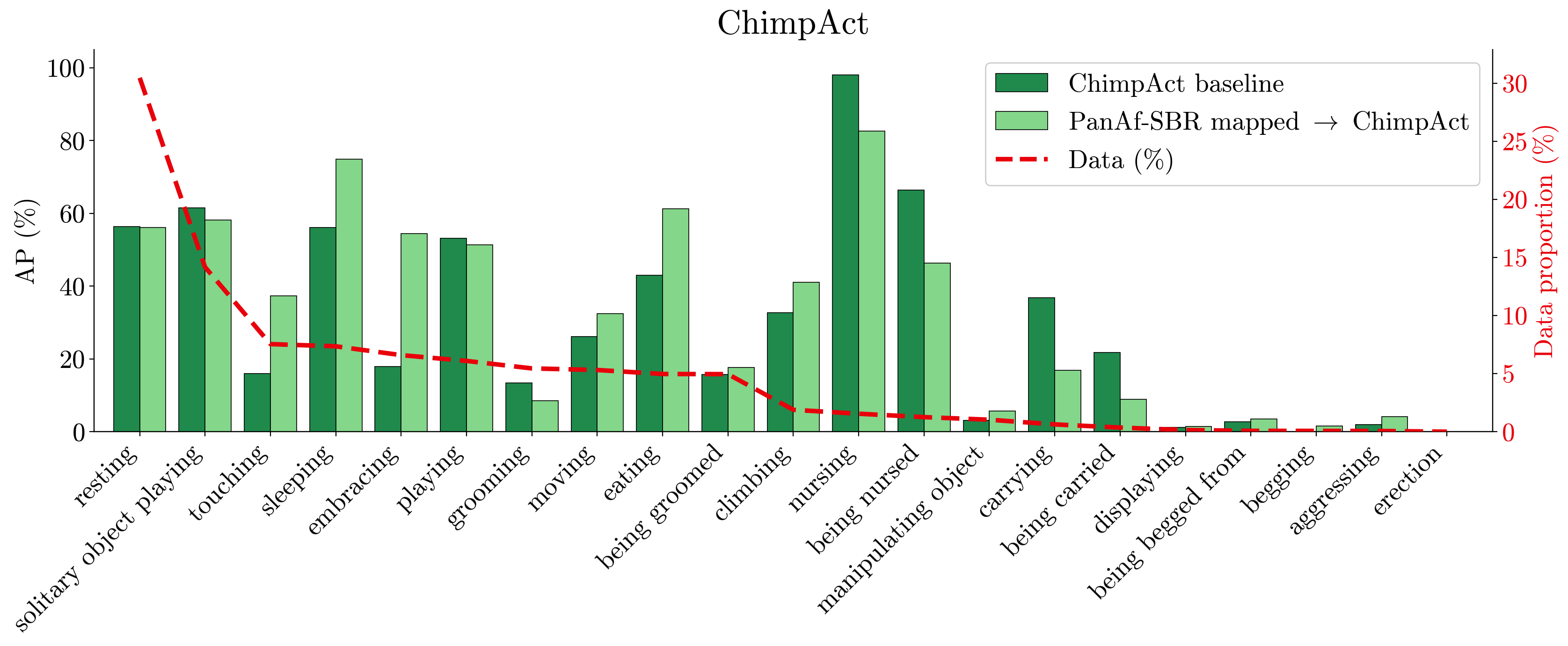}\vspace{-8pt}
\caption{\tb{\textbf{Per-class AP on ChimpACT.} We compare the replicated AlphaChimp baseline against the model pre-trained on PanAf-SBR. Locomotion classes -- \textit{sleeping}, \textit{climbing}, and \textit{moving} -- improve substantially, while the object interaction gain is driven almost exclusively by \textit{eating}. Social interaction classes show a heterogeneous response: \textit{being\_nursed} regresses by $-$20.00\%, whereas \textit{embracing} improves by $+$36.58\%.}}\vspace{-8pt}
\label{fig:per_class2}
\end{figure}

\ob{\tb{\textbf{Object Interaction Gains are Driven by \textit{`eating'}.} The class \textit{eating} improves by +18.26\% and \textit{manipulating\_object} by +2.46\%, while \textit{solitary\_object\_playing} regresses by -3.29\%. The aggregate object interaction improvement (+5.81\%) is therefore concentrated in a single class rather than distributed evenly, and the gain for \textit{eating} is somewhat unexpected given that PanAf-SBR has no eating-specific label in its own ethogram, suggesting the improvement may arise indirectly through better general object-context representations rather than direct label transfer.}}

\ob{\tb{\textbf{Social Interaction Classes show Widest Divergence.} The most severe regressions are concentrated in classes involving sustained physical contact between a caregiver and dependent: \textit{being\_nursed} ($-$20.00\%), \textit{carrying} ($-$19.88\%), \textit{nursing} ($-$15.44\%), and \textit{being\_carried} ($-$12.91\%) all degrade substantially. By contrast, \textit{embracing} improves dramatically (+36.58\%), and \textit{touching} (+21.36\%) and \textit{being\_groomed} (+1.92\%) also improve, while \textit{grooming} regresses ($-$4.84\%). This heterogeneity explains why the aggregate social interaction change is small and near-neutral ($-$0.88\% in Fig.~\ref{fig:pretrain_ablation}) despite substantial movement at the per-class level in both directions.}}

\ob{\tb{\textbf{PanAf-SBR Ethogram Labels fix Learning Horizon.} \textit{taking\_object} and \textit{losing\_object} both score 0.00\% under both configurations, and \textit{erection} is negligible in both (0.01\% vs.\ 0.04\%), consistent with these behaviours having no corresponding PanAf-SBR supervision to draw on. \textit{begging} and \textit{being\_begged\_from} likewise remain low in both configurations despite small positive deltas, suggesting these classes are fundamentally under-supported by either dataset rather than meaningfully benefiting from wild pre-training.}}

\subsection{Removing the Background from Scenes}
\label{sec:silhouette-method}

\begin{figure}[!h]
    \centering
    \includegraphics[width=0.49\textwidth]{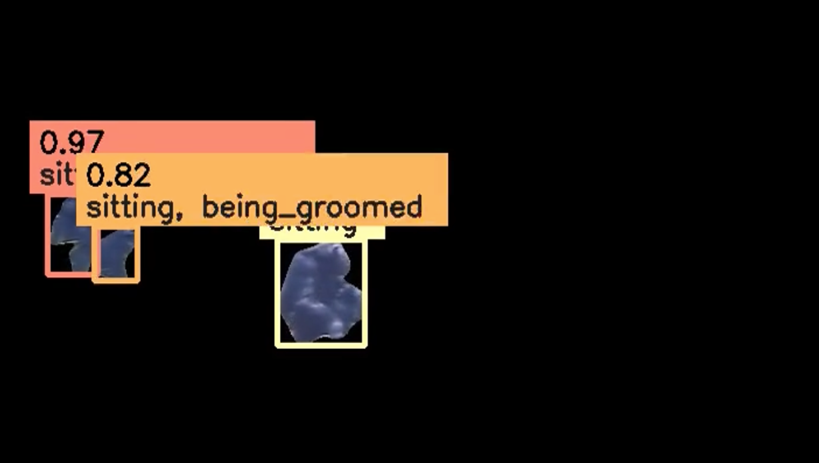}\hfill
    \includegraphics[width=0.49\textwidth]{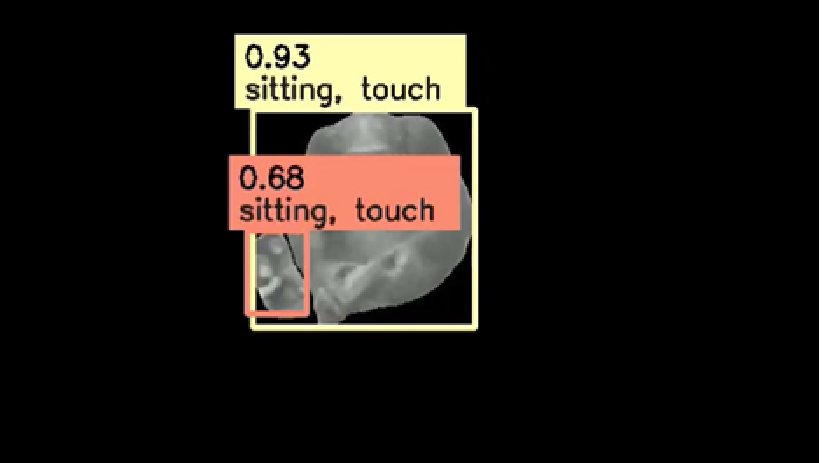}\\[0.5em]\vspace{-8pt}
    \caption{\tb{\textbf{Input Space Background Removal.} Segmentation masks are used to set backgound images to zero RGB value. Show are two `object only' frames with a superimposition of bounding boxes and behaviours exhibited.}}\vspace{-16pt}
    \label{fig:removal}
\end{figure}

\tb{\textbf{Input Space Background Deletion Procedure.} We investigated whether removing the background from scenes would improve performance. Specifically, we introduce a simple preprocessing step in which all pixels outside the segmentation masks are set to black, while pixels within the masks retain their original RGB values~(see~Fig.~\ref{fig:removal}). The resulting silhouette videos share the same resolution, frame rate, and annotations as the source videos and require no architectural changes to AlphaChimp.} 

\begin{table}[!hb]
  \scriptsize
  \centering
  \caption{\tb{
    \textbf{Effect of Background Removal on PanAf-SBR.}
    \textmd{AlphaChimp trained and evaluated on PanAf-SBR against the background-removed variant. Background removal degrades overall mAP while improving precision, recall, and F1 at the fixed confidence threshold (0.5). All scores are at IoU and classification threshold of 0.5.}}
  }
  \label{tab:silhouette}\vspace{-5pt}
  \begin{tabular}{@{}lrrrr@{}}
  \toprule
  \textbf{Configuration} & \textbf{mAP} & \textbf{Precision} & \textbf{Recall} & \textbf{F1-score} \\ \midrule
  PanAf-SBR            & 49.72 & 25.82 & 21.53 & 20.02 \\
  PanAf-SBR (Masked)   & 28.11 & 31.95 & 40.77 & 33.21 \\
  Diff ($\Delta$)     & $-$21.61 & $+$6.13 & $+$19.24 & $+$13.19 \\
  \bottomrule\vspace{-18pt}
  \end{tabular}
\end{table}

\tb{\textbf{Performance Re-alignment under Background Removal.} As shown in Table~\ref{tab:silhouette}, background removal produces a substantial drop in mAP (-21.61\%, 49.72\% to 28.11\%), indicating that background context carries significant discriminative information for behaviour recognition in wild camera trap footage. However, precision (+6.13\%), recall (+19.24\%), and F1 (+13.19\%) nevertheless improve for the chosen precision-recall trade-off (i.e., the working point).}

\tb{The per-class results in Table~\ref{tab:silhouette_grouped} reveal that AP performance decreases for nearly all classes under background removal, with only three social interaction classes improving. The only classes that benefit from background removal are social interaction classes involving direct physical contact or spatial co-occurrence between individuals: \textit{being_carried} (+17.88\%), \textit{touch} (+9.56\%), and \textit{chimp_carrying} (+1.33\%). For contact-based interactions, the relative positioning and body configuration of two individuals may be more salient when background clutter is suppressed, whereas for locomotion and posture-based classes, the background provides contextual cues the model relies upon heavily.}

\begin{table}[t]
  \scriptsize
  \centering
  \caption{\tb{
    \textbf{Per-category AP at IoU 0.5 for Background Removal on PanAf-SBR.}
    \textmd{Results are grouped into locomotion, object interaction, and social interaction categories. Background removal degrades mAP across all behaviour categories, indicating that background context is broadly discriminative for behaviour recognition in wild camera trap footage. All scores are AP (\%) at IoU threshold 0.5.}}
  }
  \label{tab:silhouette_grouped}\vspace{-5pt}
  \begin{tabular}{@{}lrrr@{}}
  \toprule
  \textbf{Configuration} & \textbf{Locomotion} & \textbf{Object Int.} & \textbf{Social Int.} \\ \midrule
  PanAf-SBR            & 55.40 & 44.07 & 44.04 \\
  PanAf-SBR (Masked)   & 31.14 &  0.14 & 28.64 \\
  Diff ($\Delta$)     & $-$24.26 & $-$43.93 & $-$15.40 \\
  \bottomrule
  \end{tabular}
\end{table}

\section{Conclusion}\vspace{-8pt}

\tb{We introduced PanAf-SBR, the first wild great ape camera trap dataset annotated with social behaviours, extending PanAf500 with 100 additional videos, 36,063 frames, and 81,096 detections spanning seven novel social behaviour classes defined under the action giver and receiver convention of ChimpACT. Using PanAf-SBR, we established a base benchmark for social behaviour recognition in wild great apes via the AlphaChimp architecture. Our bidirectional transfer learning experiments between PanAf-SBR and the captive ChimpACT dataset showed that cross-dataset pre-training is class-selective rather than uniformly beneficial or detrimental, with shared behaviour labels alone being insufficient to guarantee positive transfer. Finally, background removal revealed a clear divergence between behaviour categories -- recognition performance degrades most behaviours, yet improves recognition of physical-contact social behaviours such as touching and carrying. This work has several limitations. First, the seven social behaviours annotated here do not capture the full breadth of social interaction observed in wild great apes; behaviours such as food sharing, for example, remain unannotated and are natural candidates for future extension of the dataset's ethogram. Secondly, we note an absence of comparative baselines against alternative architectures, a gap made more difficult since AlphaChimp's design precludes direct comparison with models addressing only task subsets without substantial re-engineering; we plan to benchmark additional joint architectures on PanAf-SBR in future work. Despite these limitations, we believe PanAf-SBR provides a valuable resource: it is, to our knowledge, the only dataset enabling the study of fine-grained social behaviour recognition in wild great apes, and we hope its release will establish this as a viable area of research, supporting both conservation and study of social evolution in our closest living relatives.}

\bibliographystyle{splncs04}
\bibliography{main}

\clearpage

\end{document}